# Transfer Learning with Pretrained Remote Sensing Transformers

Anthony Fuller, *Graduate Student Member, IEEE*, Koreen Millard, *Member, IEEE,* and James R. Green, *Senior Member, IEEE*

*Abstract*—Although the remote sensing (RS) community has begun to pretrain transformers (intended to be fine-tuned on RS tasks), it is unclear how these models perform under distribution shifts. Here, we pretrain a new RS transformer—called SatViT-V2—on 1.3 million satellite-derived RS images, then fine-tune it (along with five other models) to investigate how it performs on distributions not seen during training. We split an expertly labeled land cover dataset into 14 datasets based on source biome. We train each model on each biome separately and test them on all *other* biomes. In all, this amounts to 1638 biome transfer experiments. After fine-tuning, we find that SatViT-V2 outperforms SatViT-V1 by 3.1% on in-distribution (matching biomes) and 2.8% on out-of-distribution (mismatching biomes) data. Additionally, we find that initializing fine-tuning from the linear probed solution (i.e., leveraging LPFT [1]) improves SatViT-V2's performance by another 1.2% on in-distribution and 2.4% on out-of-distribution data. Next, we find that pretrained RS transformers are better calibrated under distribution shifts than non-pretrained models and leveraging LPFT results in further improvements in model calibration. Lastly, we find that five measures of distribution shift are moderately correlated with biome transfer performance. We share code and pretrained model weights. (https://github.com/antofuller/SatViT)

## I. Introduction

Transfer learning [2] plays an essential role in remote sensing (RS) due to the scarcity of accurate reference data in relation to the large datasets required to adequately train highly complex models. This scarcity is due to the high costs of data annotation in RS, often requiring additional information gathered from fieldwork or high-resolution imagery, both of which can be prohibitively expensive or entirely unavailable (a lack of satellite coverage, physical or political barriers, etc.). Lacking reference data to train a well-performing model, RS researchers or practitioners must leverage transfer learning by transferring the knowledge (in the form of model parameters) learned on one region to another (spatial transfer) or learned on one task to another (task transfer)—we study both types of transfer learning in this article.

Now the dominant paradigm in machine learning, pretraining refers to training a model on a pretext task with the intention of fine-tuning the model on a specific downstream task of interest. Numerous deep learning studies [3]–[7] have demonstrated that scale (i.e., the amount of pretraining data, compute, and model size) contributes most to downstream performance—these scaling factors are far more important than slight modifications in model architectures, optimizers, learning rate schedules, or expertly engineered inductive biases. A new model architecture that has been demonstrated to scale incredibly well (to billions of samples and parameters [4]–[6]) is the transformer [8]. An attention-based architecture, transformers have been leveraged at scale to achieve state-of-the-art (SOTA) performance across most CV tasks [9]–[13]. Very recently, pretraining transformer models on RS data have led to SOTA performance on some RS tasks [14]–[18]. As these methods continue to develop in RS, we expect more RS tasks will be dominated by large pretrained transformers—following broad trends in deep learning.

Estimating the accuracy of a classifier without access to labeled target data is valuable to all users of RS-derived insights or data products. For example, when model predictions have real-world consequences (e.g., distributing humanitarian aid [19]), humans can consider uncertainty estimates to disregard model predictions entirely or request a second opinion [20]. In this article, we explore two types of methods that estimate the accuracy of a classifier. First, model calibration refers to how well the model's output probabilities reflect its predictive uncertainty [21]. For example, if a model was perfectly calibrated, model predictions with an 80% probability would be correct 80% of the time. Second, distribution shift refers to when the training and testing data are sampled from different distributions [22]. For example, if the training and testing data were sampled from different biomes (which we study in this article) or temporal periods.

As pretrained RS transformers become prevalent, understanding their transfer learning capabilities will be crucial to researchers and practitioners wishing to leverage these models. Following this observation, we pretrain an RS transformer (SatViT-V2) on 1.3 million paired Sentinel-1 & 2 images. Next, we fine-tune SatViT-V2 and five other baseline models on a land cover pixel classification task on 14 biomes separately and evaluate how they perform on *other* biomes. When testing under these distribution shifts (i.e., mismatching biomes), we find that:

- Pretraining improves the performance and calibration of RS transformers under biome transfer;
- Leveraging linear probe fine-tuning [1] improves the performance and calibration of pretrained RS transformers under biome transfer (this improvement in model calibration is a novel finding that can be leveraged outside of RS);

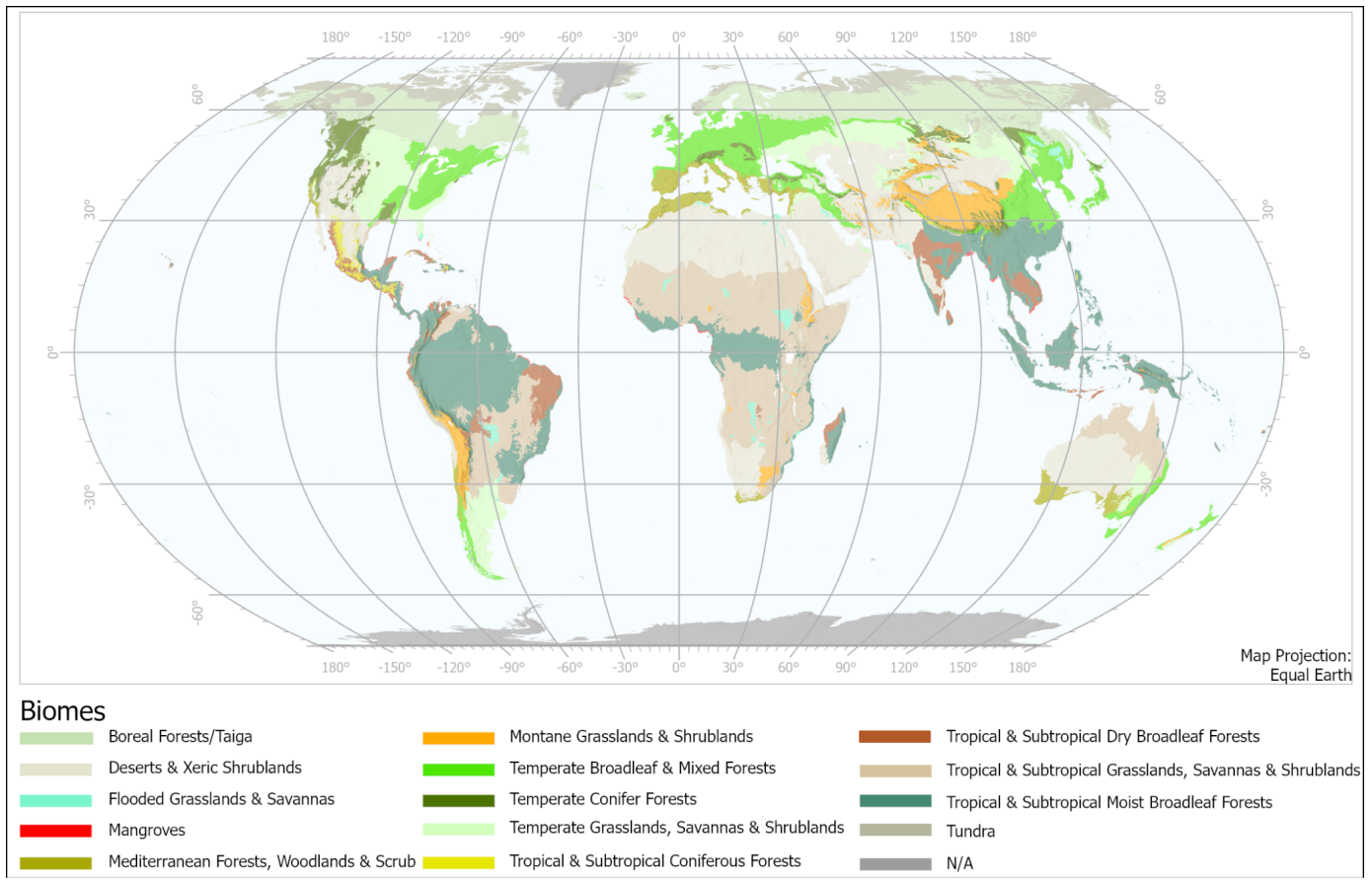

**Fig. 1.** Map of the Earth displaying the geographic boundaries of the 14 biomes that we leverage to investigate how our models trained on one biome perform on other biomes (i.e., under distribution shifts).

- Five measures of distribution shift are moderately correlated with the performance of RS transformers (i.e. transferring between similar biomes is more successful than transferring between dissimilar biomes). This paper suggests a number of indicators that can help SatViT-V2 users predict performance on target datasets.

## II. Related work

Starting with Marmanis et al. [23], RS researchers have leveraged transfer learning by initializing their models with models pretrained by the CV community on the ImageNet dataset [24]. However these models are RGB-only and were pretrained on object-centric images typical of the internet. The first *RS-specific* pretrained models employed convolutional neural networks (CNNs) with either supervised [25], [26] or self-supervised pretraining [27]–[31] to achieve higher performance than competing approaches (e.g., transferring ImageNet models or training models from scratch). Following broader trends in deep learning, Fuller et al. [18] pretrained a vision transformer [32] (ViT) on 1.3 million paired Sentinel-1 and Sentinel-2 images using a SOTA self-supervised learning algorithm called masked autoencoding [10] (MAE)—this model is called SatViT-V1. We build on this work by (i) initializing model parameters from SatViT-V1, (i) pretraining for another 120 epochs (on the same 1.3 million sample dataset), and (iii) decreasing the patch size to retain more fine-grained information—our model is called SatViT-V2.

Although not studied in the context of spatial transfer, RS researchers have proposed methods of estimating the accuracy of model predictions by comparing the predictions of an ensemble of models [33]–[35]. Outside of RS, Hendrycks and Gimpel [36] demonstrated the effectiveness of using maximum softmax probabilities to estimate the accuracy of predictions made by deep learning models. This method is straightforward and has been validated in follow-up work [22], [37]–[40]; we leverage it in section IV-D and compute model calibration based on the correlation between model confidence and model accuracy.

Numerous RS studies [41]–[50] have employed spatial transfer with deep learning models; notably, Qui et al. [51] employ spatial transfer with transformers that were pretrained by the CV community (on ImageNet data), whereas we employ spatial transfer with transformers pretrained on RS data (on Sentinel-1 and 2 data). However, none of these studies attempt to estimate model accuracy under distribution shifts. Intuitively, transferring knowledge between similar biomes should be more successful than transferring knowledge between dissimilar biomes. In CV [22], [52], researchers have validated this intuition by finding correlations between the

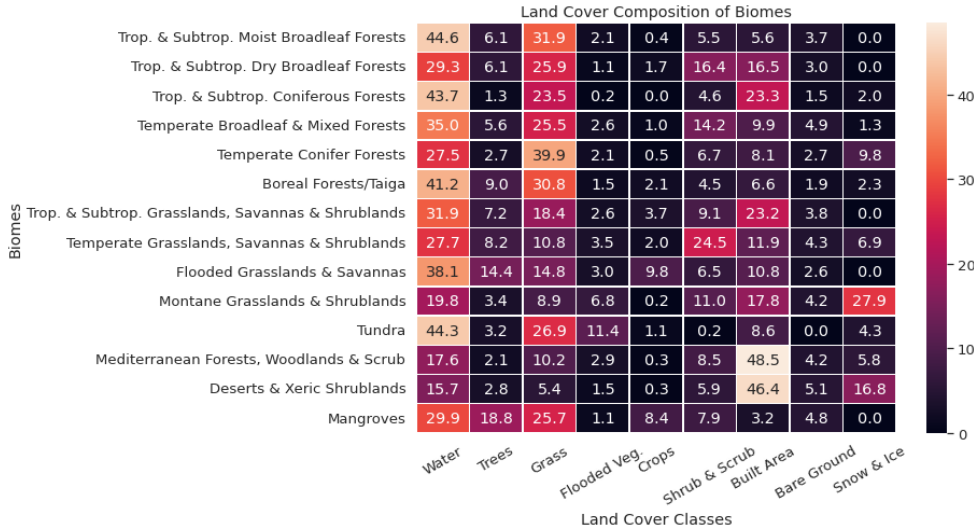

**Fig. 2.** Heatmap displaying the land cover composition of each of our 14 biomes. Composition is represented by the percent of pixels belonging to each of the 9 DynamicWorld [53] classes.

Frechet distance between the features (i.e., the pretrained representations) of datasets and their transfer performance. We leverage this method, along with other measures of distribution shift, in section IV-E.

## II. DATA COLLECTION

### A. Pretraining Data Collection

We collect a pretraining dataset of 1.3 million images sampled from six regions. The six pretraining regions are North America, South America, Europe, Sub-Saharan Africa, South-East Asia, and North-East Asia. Each region contains an equal number of samples (roughly 217 k); this is the same dataset used to pretrain SatViT-V1 [18].

All samples are paired Sentinel-1 and Sentinel-2 data downloaded via Google Earth Engine (GEE). Acquired by Sentinel-1, our synthetic aperture radar (SAR) data has a 10 m per pixel resolution in Ground Range Detected (GRD) format, all pixel values are in decibels (GEE's default), and the data were acquired in Interferometric Wide (IW) mode. Acquired by Sentinel-2, our multispectral optical data has a 10 m per pixel resolution (after reprojecting optical bands that have a coarser spatial resolution to 10 m) without atmospheric corrections (i.e., Level-1C). Our Sentinel-2 imagery consists of 13 optical bands. Our Sentinel-1 data consists of 2 SAR bands (representing backscatter from VV and VH polarizations). When a sampled location is covered by both ascending and descending Sentinel-1 orbits, we randomly choose one orbit to query data from; if only one orbit is available in the chosen month, then it is chosen automatically.

We acquire an image by first randomly sampling a location (in latitude and longitude) within our areas of interest. We then reproject the latitude and longitude coordinates to the coordinates of the local Universal Transverse Mercator (UTM) zone and create a bounding box of size 2.56 $km^2$ by 2.56 $km^2$—centered at the sampled location. Next, we query both Sentinel-1 & 2 data using our bounding box, a random year between 2016 and 2021, and a random month. Then, we take the median of all images acquired in our selected month, rejecting optical images with excess cloud cover (i.e. with greater than 30% cloud cover inside the bounding box). This leaves us with a Sentinel-1 image of shape {256, 256, 2} and a Sentinel-2 image of shape {256, 256, 13} which are stacked along the channel dimension to form a tensor of shape {256, 256, 15}—we refer to these input tensors as "images" throughout this article. We repeated this process until we reached the desired number of samples.

### B. Fine-tuning Data Collection

DynamicWorld [53] collected expert land-use land-cover (LULC) annotations of 4000 image tiles; each tile is 510 by 510 pixels at 10 m spatial resolution. The expert labelers were also given access to high-resolution imagery from Google Maps and street-view imagery from Google Street View to ensure high-quality references. Crucially, the locations of these 4000 tiles were sampled from 14 biomes [54]; this allows us to investigate how models trained on one biome perform when tested on another biome. The global geographic boundaries of our 14 biomes are visualized in Figure 1, with each color representing a different biome. Our 14 biomes are:

1) Tropical & Subtropical Moist Broadleaf Forests (291 tiles)
2) Tropical & Subtropical Dry Broadleaf Forests (283 tiles)
3) Tropical & Subtropical Coniferous Forests (214 tiles)
4) Temperate Broadleaf & Mixed Forests (304 tiles)
5) Temperate Conifer Forests (353 tiles)
6) Boreal Forests/Taiga (252 tiles)
7) Tropical & Subtropical Grasslands, Savannas & Shrublands (299 tiles)

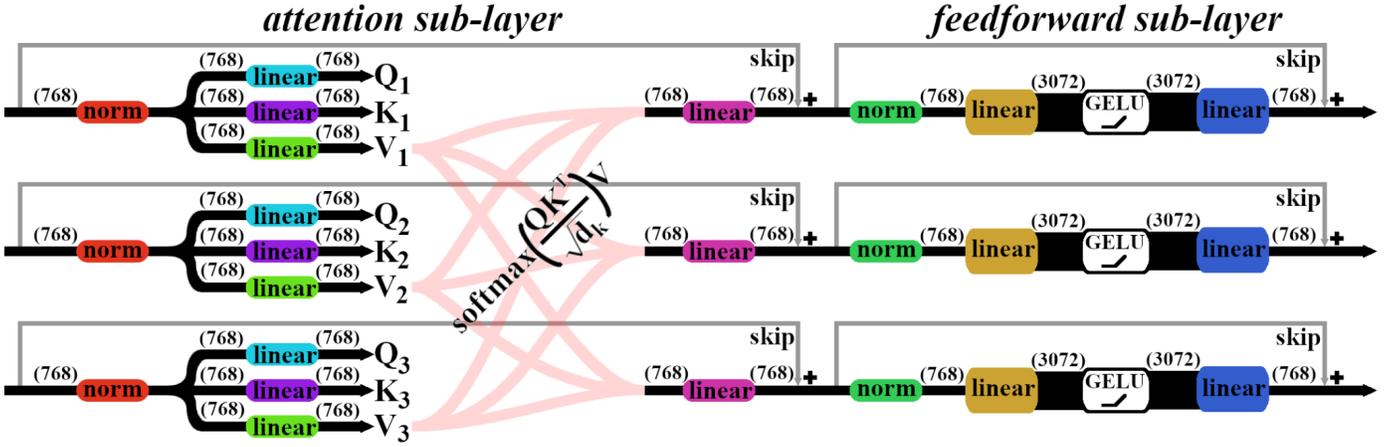

**Fig. 3.** Graphical illustration of a transformer block composed of attention and feedforward sub-layers. In this example, we display a model width of 768, a feedforward multiplier of 4 (the inner feedforward dimension is 768*4), and a single attention head. We do not depict "multi-headed" attention, which refers to splitting up queries, keys, and values into "heads" (i.e., smaller feature vectors), performing independent attention operations and then concatenating their resultant vectors. All parameters are shared across the sequence dimension, illustrated by matching colors.

8) Temperate Grasslands, Savannas & Shrublands (325 tiles)
9) Flooded Grasslands & Savannas (253 tiles)
10) Montane Grasslands & Shrublands (304 tiles)
11) Tundra (160 tiles)
12) Mediterranean Forests, Woodlands & Scrub (352 tiles)
13) Deserts & Xeric Shrublands (335 tiles)
14) Mangroves (279 tiles)

Furthermore, our 9 LULC classes, defined by DynamicWorld [53], are:
1) Water (rivers, ponds, lakes, oceans);
2) Trees (wooded vegetation, dense green shrubs, plantations, dry swamp)
3) Grass (meadows and fields without tree cover, open savanna, parks, golf courses)
4) Flooded vegetation (flooded mangroves, emergent vegetation)
5) Crops (corn, wheat, soy, hay and fallow plots of structured land)
6) Shrub & Scrub (moderate to sparse cover of bushes, shrubs, and tufts of grass)
7) Built area (cluster of houses, dense villages/towns/cityscapes, human-made surfaces)
8) Bare ground (exposed rock or soil, desert and sand dunes, empty lots in urban areas)
9) Snow & Ice (glaciers, permanent snowpack, snowfall)

Figure 2 displays the amount of annotated pixels belonging to each class as a fraction of the total annotated pixels in each biome. Water is the most frequent class in all biomes other than Montane Grasslands & Shrublands, which is dominated by snow and ice; Mediterranean Forests, Woodlands & Scrub which is dominated by built area; and Deserts & Xeric Shrublands which is also dominated by built area. Not every pixel in these datasets is annotated, so the class percentages reported in Figure 2 only represent the fraction of *annotated* pixels, not all pixels in our data. During training and testing, unlabeled pixels are ignored.

### III. METHODS

*A. Transformers*

We select a vision transformer (ViT), a transformer-based CV model, for our model architecture. Transformers receive a set of feature vectors, gradually transforming them with a series of transformer layers, each composed of two sub-layers. The number of transformer layers in a transformer is referred to as the model depth. The length of each feature vector is referred to as the model width. Finally, the number of feature vectors is referred to as the sequence length. The architecture we describe is often called a "vanilla" transformer and is the most common transformer-based architecture in deep learning.

As depicted in Figure 3, the first sublayer uses multi-headed self-attention (MHSA) [8] to exchange features across the sequence (i.e., between feature vectors). MHSA first normalizes (using a LayerNorm [55]) each feature vector, then uses three linear projections to create query, key, and value vectors. Queries and keys are then compared (via their dot products) to build an attention matrix (populated by attention scores), which is then divided by a scalar. This scalar is the square root of the length of key feature vectors. Next, a softmax is applied to the attention matrix; now, attention scores will sum to 1 along the key axis of the attention matrix. In the final attention operation, new feature vectors are built by taking the weighted sum of value vectors, where the weights are taken from the normalized attention matrix. In other words, value vectors are routed across the sequence based on the relative similarity between queries and keys. Not depicted in Figure 3, is the "multi-headed" component of MHSA. This refers to splitting up queries, keys, and values into "heads" (i.e., smaller feature vectors), performing independent attention operations, and then concatenating their resultant vectors. Finally, another linear projection is performed, and a skip connection is added.

The second sub-layer uses a feedforward network (FFN) to non-linearly transform the features of individual feature vectors—no information is exchanged across the sequence in this sub-layer. FFN first normalizes (also using a LayerNorm) each feature vector, then uses a linear projection that increases the length of feature vectors (typically by a factor of 4). Next, a GELU activation function [56] is applied, then another linear layer projects the feature vectors back down to their original sizes. Finally, another skip connection is added.

Figure 3 depicts a single transformer block with a model width of 768, a sequence length of 3, and an FFN inner dimension of 3072. Crucially, all model parameters are shared across the sequence dimension (matching colors indicate weight sharing).

*B. Vision Transformers*

Adapting the transformer architecture to images requires processing images into a sequence of feature vectors, which are then processed by a vanilla transformer described in section 3-A. First, we split an image into "patches"; these are typically squares of height and width of 8 or 16 pixels. For SatViT-V2, we use 8 by 8 patches (along with 15 channels); thus, each patch is a tensor of shape {8, 8, 15}. This deviates from SatViT-V1, they use 16 by 16 (along with 15 channels); thus, each patch was a tensor of shape {16, 16, 15}. Next, we flatten these 3-dimensional volumes resulting in vectors of length 960. Finally, we linearly project these 960-d vectors to our encoder's model width, i.e., 768 features. Like the transformer architecture (Figure 3), these linear projection layers also share weights across the sequence dimension; in other words, every patch uses the same linear transformation to project input pixels to patch embeddings. Given that our images are 256 by 256 pixels, and our patches are 8 by 8 pixels, patchification results in 1024 total patch embeddings. Before sending patches to our transformer model, we add a 2D-sinusoidal position embedding to each patch embedding; these position embeddings represent the patch's location within the image.

We use the ViT-Base [32] architecture for our SatViT-V2 encoder, which has 12 transformer layers (model depth of 12), 768 features per patch (model width of 768), an inner FFN dimension of 3072, and 12 attention heads; this is identical to SatViT-V1, except for the smaller patch size. We chose a smaller patch size for SatViT-V2 because preliminary results indicated that SatViT-V1 did not perform well on dense pixel classification tasks, especially those with images containing fine-grained input detail.

*C. Masked Autoencoding (MAE) Pretraining*

In order to pretrain our ViT encoder, we introduce a decoder and employ masked autoencoding (MAE) [10]—a state-of-the-art self-supervised learning algorithm. In short, MAE hides a high proportion of patches, then trains an encoder and decoder to reconstruct the hidden patches. In more detail, after patchifying the images, we randomly mask out 75% of patches. After our encoder processes the visible patches, we send the patch encodings to our decoder. Our decoder linearly projects the 256 (25% of 1024 patches) visible patch encodings to 512 features per patch (our decoder model width). Mask embeddings are then introduced as placeholders representing the original hidden patches. All 768 (75% of 1024 patches) mask embeddings share the same 512 features. Next, the patch encodings and mask embeddings are concatenated along the sequence dimension, and 1024 (one for each patch) 2D-sinusoidal position embeddings are added. The decoder (with only 1 transformer layer) outputs feature vectors of length 512 (our decoder model width) for all 1024 patches. Finally, the feature vectors representing the original hidden patches are linearly projected to 960 values (the total number of pixels per patch) and reshaped to tensors of shape {8, 8, 15} to form patch predictions. Our self-supervised learning objective minimizes the mean squared error between these patch predictions and our held-out patches.

We pretrain SatViT-V2 for 120 epochs on our 1.3 million sample dataset of stacked Sentinel-1 & 2 images. We use the AdamW optimizer [57], a batch size of 256, a max learning rate of 1.5e-4, and a cosine learning rate schedule. We perform channel-wise data normalization and augmentation by rotating and flipping images. Importantly, we initialize our encoder's weights with SatViT-V1's pretrained weights; since SatViT-V1 was pretrained for 100 epochs, this means SatViT-V2 has been pretrained on all 1.3 million images 220 total times. The only encoder weights we cannot initialize from SatViT-V1 are the patch projections since SatViT-V2 uses smaller patches. Therefore, these patch projections and SatViT-V2's decoder are randomly initialized.

Our pretraining dataset consumes 2.6 TB of storage; this is far larger than our machine's memory. As a result, we need to load batches of data from storage while pretraining our model. This slows down pretraining significantly, even when loading data asynchronously. In order to improve GPU utilization, we employ a batch repetition [58] factor of 8; this means that each time we load a batch of data, it is sampled 8 times before moving on to the next batch of data. On our system (RTX 3090 GPU and data stored on SSDs), this speeds-up pretraining by around 6 times. This method has recently been leveraged in MAE pretraining on video data which also suffers from file IO bottlenecks [59], [60]. As a result, each image is sampled 120 times (while pretraining SatViT-V2) but only loaded from storage 15 times.

*D. Transfer Learning on DynamicWorld*

This article investigates how models trained on one biome perform when tested on a different biome—this is a particular case of spatial transfer. To this end, we split each of our 14 datasets (one for each biome) into a training (80%) and validation (20%) split. First, we fine-tune our model on the training set for 100 epochs for each biome. After each epoch, we calculate the overall accuracy on the validation set and save the model parameters associated with the best validation accuracy; this leaves us with 14 fine-tuned SatViT-V2 models (one for each biome). Then, we test each model on the *other* 13 biomes. We refer to the overall accuracy on the validation set as in-distribution (ID) accuracy since the validation images are sampled from the same biome as the training images. We refer to the overall accuracy on the test sets as out-of-distribution (OOD) accuracy since test images are *not* sampled from the same biome as the training images. Additionally, since DynamicWorld annotations are on tiles of 510 by 510 pixels, but SatViT-V1 and SatViT-V2 receive images of shape 256 by 256 pixels, we split each tile into 4 images. This means images cut from the same tile share identical 2-pixel strips (since our tiles are not 512 by 512). To ensure these small strips of pixels are not shared across our train and validation splits, we sample our splits based on tiles, not images.

To fine-tune our model, we initialize our encoder with SatViT-V2's pretrained encoder weights (leaving the decoder unused). Next, we add a linear "head" (i.e., a dense/fully-connected layer) after the final ViT encoder layer. This linear head—which is randomly initialized—projects each patch encoding (768 features encode each patch) to the number of pixels per patch (8 by 8) times the number of classes (9 in our case); this leaves us with 9 logits per pixel in our image. Crucially, unlike pretraining, the entire image is encoded (i.e., no masking is applied) during fine-tuning and testing. During fine-tuning, we update the parameters of both the encoder and linear head according to the cross entropy loss between DynamicWorld annotations and our model's predictions.

We also perform a "linear probe" experiment which is set up exactly like fine-tuning, except our encoder's pretrained parameters are not updated. Instead, it relies on the linear head to separate the classes given patch encodings.

Across deep learning, there is typically a linear relationship between ID and OOD accuracy [61]. Furthermore, models that perform better OOD than their ID accuracy would predict, exhibit effective robustness (ER). Andreassen et al. [62] found that when fine-tuning pretrained models, they exhibit ER early in fine-tuning but their ER vanishes at convergence. Building on this finding, Kumar et al. [1] found that fine-tuning can *distort* a pretrained model's representations so much that it underperforms on OOD test sets, relative to a linear probe. To address this, they propose linear-probe fine-tuning (LPFT). LPFT first trains a linear probe, then adds another stage of training by fine-tuning both the pretrained model and the linear head's parameters. Crucially, when fine-tuning, the linear head is initialized from the linear probed solution. They show LPFT results in better OOD performance without harming ID performance.

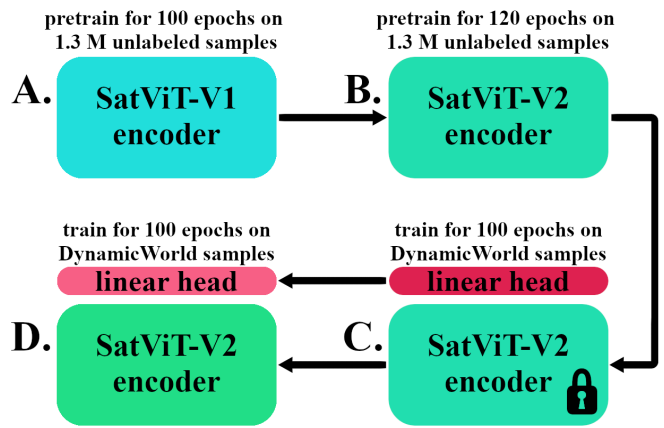

**Fig. 4.** Model training stages. Changes in color represent updated model parameters. The "lock" icon indicates that model parameters are not updated during this stage of training.

Figure 4 depicts how model parameters are transferred between stages of training. LPFT incorporates all stages, A through D. Fine-tuning skips stage C by directly fine-tuning our pretrained model without the linear probing stage. Finally, the linear probe experiment skips stage D by never fine-tuning our pretrained model's parameters.

Finally, to put SatViT-V2's performance in context, we also fine-tune SatViT-V1 [18], four convolutional neural networks trained from scratch (UNet [63], UNet++ [64], ResUNet [65], and ResUNet++ [66]), and SatViT-V2 trained from scratch. This allows us to compare the transfer learning performance of pretrained RS transformers with models that are not pretrained. Enabling us to perform many experiments, we do not perform extensive hyper-parameter tuning. Instead, we select reasonable hyper-parameters (batch size of 32, learning rate of 1e-4, flipping and rotating for data augmentation, and the AdamW optimizer) and validate our model after every training epoch to perform early stopping.

*E. Model Calibration under Distribution Shift*

To investigate model calibration under distribution shifts, we gather estimates of model confidence. We define model confidence as the probability of the predicted class (i.e. the maximum probability). In this article we study seven estimates of model confidence via temperature scaling.

To calculate these probabilities, we calculate the softmax over the logits (the outputs of our model's final layer) of our 9 LULC classes. However, since applying a temperature to the softmax calculation can improve calibration [38], we sweep over seven temperatures {0.1, 0.25, 0.5, 1.0 1.5, 2.0, 3.0} to investigate its effect on calibration under distribution shifts. Applying a temperature to the softmax calculation is accomplished by dividing logits by a scalar value (the temperature) before the softmax. Finally, we compute correlations between model confidence and model accuracy. This means that for a model trained on biome X, we find the correlation between the accuracy on biome Y and the model's confidence in the predictions made on biome Y.

*F. Measuring Distribution Shift*

To test the hypothesis that transferring knowledge between similar biomes should be more successful than transferring knowledge between dissimilar biomes, we gather estimates of the similarity between biomes (i.e., measures of distribution shift). In this article, we study five estimates of distribution shift; four do not require target labels (i.e., annotations of the dataset we are testing on), whereas one does require target labels.

Not requiring target labels, we compare spectral and feature similarity between biomes. To calculate spectral similarity, we compute the average of all pixels each the biome, leaving us with 15-dimensional vectors (i.e., one for each band), then compare biomes by computing the cosine similarity and Frechet similarity (defined by the inverse of the Frechet distance). To calculate feature similarity, we leverage our pretrained SatViT-V2 encoder to create 768-dimensional image embeddings (image embeddings are computed by averaging the patch embeddings of each image), then compare biomes by computing the cosine similarity and Frechet similarity.

Requiring target labels, we compare the class similarity between biomes. For each biome, we create a composition vector by computing the fraction of pixels belonging to each class (these are equivalent to the rows in Figure 2), leaving us with 9-dimensional vectors (one for each LULC class), then compare biomes by computing the cosine similarity between composition vectors.

We use Spearman's Rank correlation to compute the strength of monotonic relationship between biome similarity and biome transfer performance. This means that for a model trained on biome X, we find the correlation between the accuracy on biome Y and the distribution shift between biomes X and Y.

## IV. RESULTS AND DISCUSSION

*A. In-distribution versus Out-of-distribution Accuracy*

Pretrained RS models outperform non-pretrained RS models on both ID and OOD data (Figure 5). This finding is well-established in deep learning [67], [68], thus is not surprising. Interestingly, the performance gap between pretrained and non-pretrained models is greater for OOD (by 6.5% on average) than ID data (by 3.3% on average).

After fine-tuning, SatViT-V2 outperforms SatViT-V1 by 3.1% on ID and 2.8% on OOD data. We attribute this to the additional pretraining steps (i.e., another 120 epochs of MAE pretraining) and the smaller patch sizes. However, due to the high pretraining costs, we do not run separate pretraining experiments to disentangle the two effects.

Qualitatively, we find a roughly linear relationship between the ID and OOD accuracy of our 9 models. Although this linear relationship is often found, it is not guaranteed; recent work has found no or even a negative relationship between ID and OOD accuracy on some datasets [69].

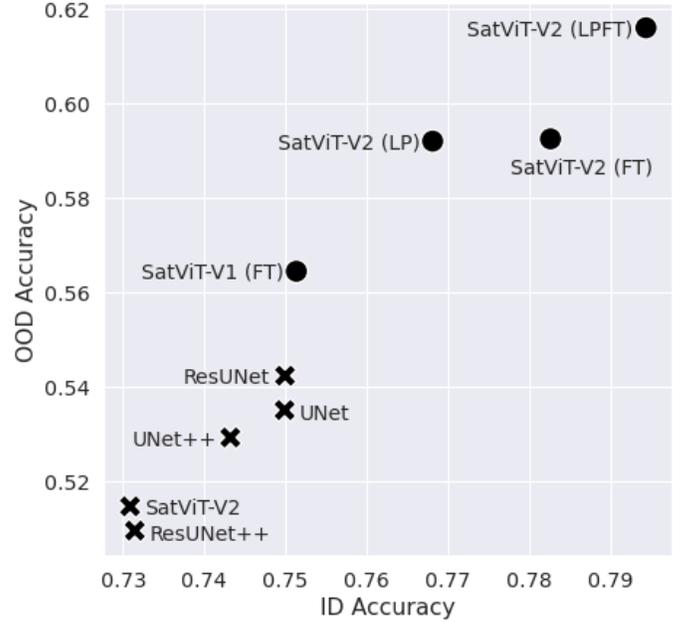

**Fig. 5.** In-distribution overall accuracy versus out-of-distribution overall accuracy. Models marked by an 'x' were not pretrained; models marked by a '•' were pretrained. Each model's ID accuracy is averaged over our 14 validation sets (on matching biomes). Each model's OOD accuracy is averaged over 182 test runs (on mismatching biomes).

Interestingly, a linear probe of SatViT-V2's encodings performs on par with fine-tuning when tested on OOD data. In this setting, our encoder's parameters (comprising 99.5% of our total model parameters) are never adjusted to our classification task and indicates that our SatViT-V2 encoder has learned general representations during pretraining. Nonetheless, fine-tuning outperforms a linear probe on ID data; this is not surprising as all 90 M model parameters are adjusted on training data (that is by definition ID). Fine-tuning outperforming a linear probe on ID data is also widely observed in deep learning [70]. Following Kumar et al.[1], we find that LPFT improves ID performance by 1.2% and OOD performance by 2.4% over fine-tuning SatViT-V2..

Finally, we find that SatViT-V2 models trained from scratch—i.e., they did not leverage pretraining—performed worse than all four convolutional models on ID data and worse than three of four convolutional models on OOD data (Figure 5). This finding agrees with prior work in computer vision that demonstrates that ViT's require large-scale pretraining [32] or higher-order optimizers [71] to overcome their weaker inductive biases (compared with convolutional models).

*B. Biome Transfer with SatViT-V2 (LPFT)*

We use our best model (our pretrained SatViT-V2 encoder that leverages LPFT) to analyze biome transfer performance by considering all combinations of source and target biomes (Figure 6).

"Source" refers to the biome on which the model was trained and validated. "Target" refers to the biome on which the model was tested (using all available images in target biomes). We believe this setup (train and validate on ID data and test on OOD data) is a realistic scenario that researchers and practitioners encounter when facing distribution shifts. When used as a source, biomes Temperate Broadleaf & Mixed Forest (67.7%), Temperate Grasslands, Savannas & Shrublands (65.8%) and Temperate Conifer Forests (65.8%) performed best when transferred to OOD data. Temperate climates and significant vegetation characterize these biomes. Conversely, biomes Tundra (43.1%), Mediterranean Forests, Woodlands & Scrub (57.1%), and Montane Grasslands & Shrublands (59.4%) performed worst when transferred to OOD data. Tundra has the least amount of training data (only 160 tiles) which likely contributes to its poor transfer performance.

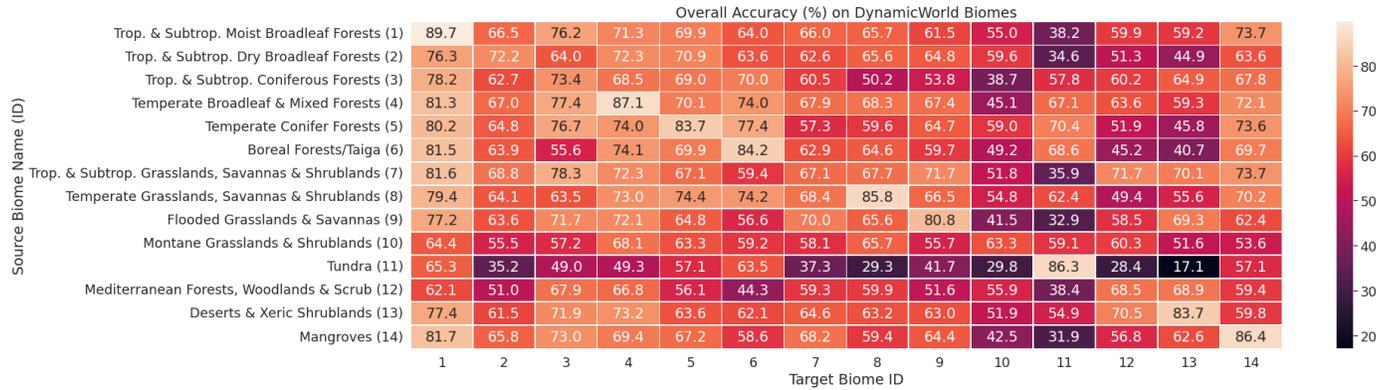

**Fig. 6.** Heatmap displaying the overall accuracy of SatViT-V2 (leveraging LPFT) on target biomes. We train our model on each source biome separately (resulting in 14 models). Then, we test all models on all target biomes. We consider matching biomes to be in-distribution (along the diagonal), and mismatching biomes to be out-of-distribution (not along the diagonal).

When used as a target, biomes Tropical & Subtropical Moist Broadleaf Forests (75.9%), Temperate Broadleaf & Mixed Forests (69.6%), and Tropical & Subtropical Coniferous Forests (67.9%) were easiest to classify for models trained on different biomes. This could indicate that the land cover classes present in forested biomes are easiest to separate. Conversely, biomes Montane Grasslands & Shrublands (48.8%), Tundra (50.2%), and Deserts & Xeric Shrublands (54.6%) were the most difficult to classify for models trained on different biomes.

Interestingly, there are three target biomes whose matching biomes do not outperform mismatching biomes; i.e., when testing on a target biome it is better to train on a different biome. These biomes are: (i) Tropical & Subtropical Coniferous Forests, (ii) Tropical & Subtropical Grasslands, Savannas & Shrublands, and (iii) Mediterranean Forests, Woodlands & Scrub.

*C. Pretrained versus Non-pretrained models*

Although the accuracy of our four pretrained models is higher than our five non-pretrained models—by 6.5%, averaged across all 14 biomes—some source biomes benefit more than others when leveraging pretrained models. For example, our pretrained models outperform non-pretrained models by 11.9% when trained on biome Flooded Grasslands & Savannas and tested on OOD data. Whereas our pretrained models only outperform non-pretrained models by 1.5% when trained on biome Tundra. Similarly, some target biomes benefit more than others when leveraging pretrained models. For example, our pretrained models outperform non-pretrained models by 9.2% when tested on biome Temperate Broadleaf & Mixed Forests. On the other hand, we see our smallest difference (3.9%) between pretrained and non-pretrained models when testing on biome Montane Grasslands & Shrublands.

*D. Model Calibration under Biome Transfer*

Figure 7 displays the Pearson correlation coefficients between OOD accuracy (averaged across our 182 combinations of source and target biomes) and model confidence—which is the probability a model gives to the class it predicts. In general, we find that models that perform better on OOD data are also better calibrated to OOD data (Figure 7). ResUNet++ performs worst on OOD data and is the worst calibrated (r = 0.223). Conversely, SatViT-V2 (LPFT) performs best on OOD data and is the best calibrated (r = 0.752). Notably, leveraging LPFT improves the calibration over vanilla fine-tuning (increase in r by 0.106). This finding—that LPFT improves the calibration of models—is both novel and potentially impactful for all applications that would benefit from well-calibrated models. Interestingly, we find that SatViT-V2 trained from scratch was better calibrated (r = 0.553) than all other non-pretrained models, despite its poor performance on OOD data (Figure 5). This aligns with prior work in CV [21], which demonstrated that ViT models are better calibrated than convolutional models, even after considering model size and pretraining amount. Finally, we find that temperature scaling is important to improve the calibration of most models. For some models (i.e., SatViT-V2 trained from scratch and SatViT-V1 fine-tuned), scaling the temperature to values greater than 1 improves calibration. For other models (ResUNet, ResUNet++, and all pretrained SatViT-V2 models), scaling the temperature to values lower than 1 improves calibration. Crucially, one would need target labels to optimally scale T to improve the calibration of models under distribution shifts. However, our best model (SatViT-V2 with LPFT) would see a negligible improvement in calibration (increase in r by 0.002) if target labels were known, and thus could be leveraged in scaling T. This is the difference between the scaled temperature of T=0.5 (r = 0.752) and the unscaled temperature of T=1 (r = 0.750).

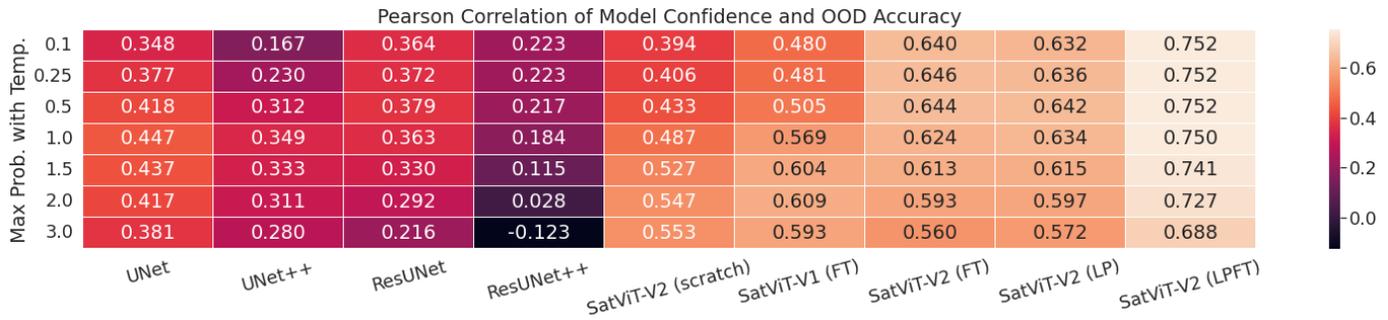

**Fig. 7.** Heatmap displaying the Pearson correlation coefficients between OOD accuracy (models trained on biome X and tested on biome Y) and model confidence (on biome Y predictions). We scale probabilities using 7 temperatures (x-axis).

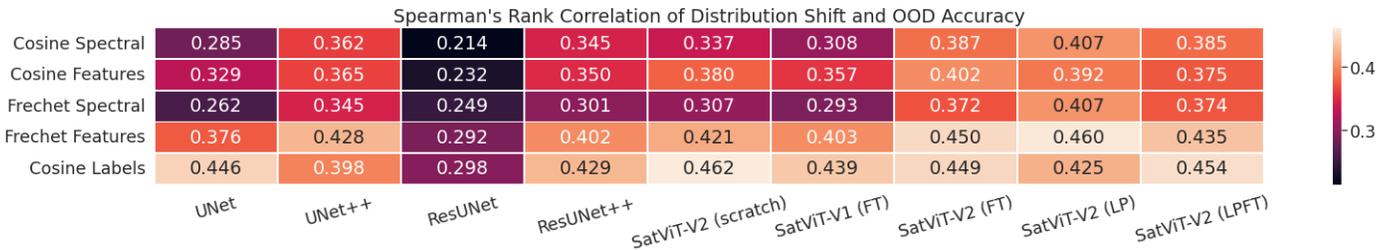

**Fig. 8.** Heatmap displaying the Spearman's rank correlation coefficients between OOD accuracy (models trained on biome X and tested on biome Y) and distribution shift (between biome X and biome Y). We estimate the amount of distribution shift using 5 measures. "Cosine" refers to the cosine similarity and Frechet refers to the Frechet similarity (inverse of Frechet distance).

*E. Distribution Shift and Biome Transfer Performance*

Across all measures, we find the amount of distribution shift (between source and target biomes) to be correlated with biome transfer performance (Figure 8). This agrees with our intuition that more similar biomes should transfer more successfully than dissimilar biomes and agrees with findings from CV [22].

We find that our pretrained SatViT-V2 models (LP, FT, and LPFT) demonstrate more predictable (and interpretable) performance under distribution shifts. Although using SatViT-V2's features to calculate the amount of distribution shift may give our SatViT-V2 models an advantage, this pattern holds even when considering spectral differences between biomes which do not rely on SatViT-V2's features. Averaged across all models, we find the similarity between composition vectors to correlate best with biome transfer performance (rho = 0.422)—this method requires annotated target data to compute land cover composition vectors. Among methods not requiring annotated target data, the Frechet similarity between the features of biomes is best correlated with biome transfer performance (rho = 0.408). This aligns with prior work in CV [22], [52] demonstrating the effectiveness of using the Frechet similarity as a measure for distribution shift to predict performance. Despite this, SatViT-V2's (with LPFT) biome transfer performance can best be explained by the confidence in its predictions; the Frechet similarity between biomes can explain 19% of the variance, whereas the model's unscaled (i.e., T=1) maximum probabilities can explain 56% of the variance.

Due to the prevalence and challenge of distribution shifts in RS data, many researchers [72]–[75] have leveraged domain adaptation to improve accuracy on target datasets, helping their models transfer to new distributions. Domain adaptation refers to adapting classifiers to improve prediction accuracy over target datasets [72]. Although an effective solution in many circumstances, domain adaptation requires resources that may be unavailable. For instance, to perform domain adaptation, one must collect unlabelled data from the target distribution, adapt the classifier, and make predictions. This delay may be unacceptable in real-time systems requiring immediate predictions, such as natural disaster monitoring. Furthermore, the technical expertise and computational resources required to adapt large deep learning models may be unavailable for many users of RS data. In this article, we study the effects of directly using a classifier on a new distribution, *without* adapting it. When domain adaptation is an available tool for users of RS data, we believe it can be leveraged on top of our findings to further improve performance under distribution shifts. We leave this investigation to future work.

## V. Conclusion

Our work is the first to investigate: (i) the performance (accuracy and model calibration) of pretrained RS transformers under distribution shifts, (ii) the performance (accuracy and model calibration) of RS transformers leveraging linear probe fine-tuning, and (iii) the correlations between accuracy and distribution shift in RS. We believe our findings will continue to increase in importance as the RS community leverages more SOTA methods from deep learning—specifically the family of models we investigate in this article (i.e., pretrained RS transformers).

We first demonstrate that our new pretrained model—SatViT-V2—significantly outperforms SatViT-V1 under biome transfer. Next, we validate the efficacy of linear probe fine-tuning by improving OOD accuracy by 2.4% relative to vanilla fine-tuning; notably, with implications beyond RS, this method also improves the calibration of models. Finally, we find moderate correlations between five measures of distribution shift and biome transfer performance, providing SatViT-V2 users with some indication of how pretrained models will perform on their respective datasets.